# Hệ Thống Robot Tuần Tra Dựa Trên Logic Mờ Ứng Dụng Trong Hệ Thống Tự Động Hóa Tòa Nhà

# Development of a Fuzzy-based Patrol Robot Using in Building Automation System

Nguyễn Thị Thanh Vân, Phùng Mạnh Dương, Phạm Đình Tuân, Trần Quang Vinh

Bộ môn Điện tử và Kỹ thuật Máy tính

Trường Đại học Công Nghệ

Đại học Quốc Gia Hà Nội

Nhà G2, 144 Xuân Thủy, Cầu Giấy, Hà Nội. Tel 043-37549272, Email: vanntt@vnu.edu.vn

*Tóm tắt*—Hệ thống tự động hoá toà nhà (BAS) cho phép giám sát và điều khiển hoạt động của các phân hệ trong tòa nhà như: điều khiển thông số môi trường, quản lý điện năng, điều khiển đóng ngắt các thiết bị điện, báo cháy, kiểm soát an ninh và các vấn đề vào ra toà nhà, v.v…Trong hệ thống tòa nhà thông minh này, hầu hết các hoạt động đều được điều khiển và theo dõi tại trung tâm, vì thế vấn đề an ninh của toà nhà phải được đảm bảo tuyệt đối. Với hệ thống giám sát truyền thống thì camera thường được lắp đặt tại các vị trí cố định do đó hạn chế phạm vi quan sát. Để khắc phục vấn đề này, bài báo trình bày một hệ thống theo dõi an ninh tại các tầng của toà nhà sử dụng robot di động như một người tuần tra. Thuật toán logic mờ được ứng dụng để giải quyết vấn đề tránh vật trong môi trường hoạt động không biết trước của robot tuần tra. Hệ thống tuần tra tự động này giúp tăng tính linh hoạt và giảm đáng kể chi phí đầu tư so với hệ thống truyền thống. Các kết quả tính toán và thực nghiệm cho thấy hệ thống thoả mãn được yêu cầu đề ra đối với nhiệm vụ giám sát an ninh tại một tầng của toà nhà.

*Abstract*: A Building Automation System (BAS) has functions of monitoring and controlling the operation of all building sub-systems such as HVAC (Heating-Ventilation, Air-conditioning Control), electric consumption management, fire alarm control, security and access control, and appliance switching control. In the BAS, almost operations are automatically performed at the control centre, the building security therefore must be strictly protected. In the traditional system, the security is usually ensured by a number of cameras installed at fixed positions and it may results in a limited vision. To overcome this disadvantage, our paper presents a novel security system in which a mobile robot is used as a patrol. The robot is equipped with fuzzy-based algorithms to allow it to avoid the obstacles in an unknown environment as well as other necessary mechanisms demanded for its patrol mission. The experiment results show that the system satisfies the requirements for the objective of monitoring and securing the building.

*Từ khoá: BAS; tòa nhà thông minh; robot an ninh; robot tuần tra; robot di động; logic mờ; tránh vật*

## I. Giới Thiệu

Thông thường, trong một hệ thống tự động hóa tòa nhà (BAS- Building Automatic System), vấn đề đảm bảo an ninh của tòa nhà được phân rõ cho một phân hệ riêng, đó là phân hệ đảm bảo an ninh và điều khiển vào ra. Phân hệ này có chức năng giám sát và điều khiển việc ra vào ở cửa trước toà nhà, phát hiện và báo động những xâm nhập bất hợp pháp. Các cửa ra vào và cửa sổ cần giám sát được lắp đặt các cảm biến tiếp xúc từ tính, các cảm biến phát hiện vỡ cửa kính, do đó có thể báo động được các xâm nhập bất thường. Các camera được lắp đặt ở các khu vực cần thiết cho phép quan sát được các hoạt động xảy ra quanh toà nhà. Đồng thời, hệ thống cũng cho phép nhận dạng và quản lý tiểu sử của mỗi nhân viên thường ra vào toà nhà bằng thẻ nhận dạng vô tuyến RFID [1]. Trong trường hợp yêu cầu đặc biệt thì camera sẽ được lắp đặt thêm tại hành lang các tầng, thậm chí là đặt ở trong phòng. Dữ liệu từ các cảm biến, camera và thiết bị báo động được cập nhật liên tục để ghi lại những hoạt động của toà nhà.

Tính an ninh của tòa nhà càng cao khi các thiết bị lắp đặt có thể bao quát hết toàn bộ hoạt động của tòa nhà. Vì phạm vi hoạt động của các camera bị hạn chế, nên muốn mở rộng phạm vi quan sát chỉ bằng cách là tăng số lượng camera. Điều này khiến chi phí mua sắm, vận hành và bảo dưỡng các thiết bị tăng lên, đồng thời hệ thống cũng trở lên phức tạp khó quản lý. Để khắc phục vấn đề này, chúng tôi đề xuất một giải pháp mới là sử dụng robot di động thực hiện chức năng giám sát tại các tầng của tòa nhà.

Trong hệ thống, robot đóng vai trò như một người tuần tra, hoạt động tự quản trị, thu thập và xử lý thông tin, đưa ra quyết định kịp thời để đảm bảo vấn đề an ninh. Robot di động được trang bị camera thường, hoặc camera toàn phương (Omni-directional Camera) có khả năng bao quát hết phạm vi của một tầng, yêu cầu hoạt động tại các thời điểm nhất định, đặc biệt là ở những giờ "giới nghiêm". Tùy vào yêu cầu của tòa nhà, hoạt động tuần tra của robot có thể có nhiều phương án, ví dụ 10h đêm đi tuần một lần, 2h sáng đi tuần lần tiếp theo, hoặc đi liên tục nhiều vòng hành lang. Quãng đường đi tuần được tính từ điểm xuất phát, di chuyển dọc theo tường của các căn hộ cho đến hết hành lang của tầng thì kết thúc. Thông thường robot sẽ di chuyển với một tốc độ ổn định dọc theo tường áp dụng thuật toán bám tường wall-following [2]. Tuy nhiên, trong môi trường hoạt động của robot di động những khó khăn thường xuất phát từ môi trường làm việc như: thông tin về môi trường không đầy đủ và mang tính tương đối, môi trường hoạt động phức tạp thay đổi không dự đoán được. Logic mờ được ứng dụng một cách hiệu quả khi xử lý các thông tin mang tính "không chắc chắn" trong các hệ thống dẫn đường và tránh vật cản của robot di động [3-7]. Trong bài báo này, chúng tôi tập trung xây dựng một hệ thống robot di động thông minh làm nhiệm vụ tuần tra và giám sát hoạt động của một tầng trong tòa nhà bao gồm việc gửi dữ liệu từ camera về trung tâm để lưu trữ, sử dụng logic mờ trong quá trình đi tuần tra để tránh vật cản, phát hiện có sự cố bất thường hay khả nghi từ đó kích hoạt hệ thống báo động của tòa nhà. Nhiều thí nghiệm đã được tiến hành để đánh giá hiệu quả cũng như khả năng ứng dụng giải pháp đề xuất.

Bài báo có cấu trúc gồm năm phần được tổ chức một cách logic như sau. Phần I giới thiệu và đặt vấn đề. Phần II mô tả cấu trúc phần cứng và phần mềm của hệ thống. Trong phần III, việc tính toán cài đặt giải thuật logic mờ cho robot được phân tích chi tiết. Phần IV và phần V trình bày thực nghiệm và kết luận.

## II. CẤU TRÚC HỆ THỐNG

Trong bài báo này, chúng tôi đã xây dựng một hệ thống an ninh tòa nhà sử dụng robot di động thông minh bao gồm cấu trúc phần cứng và các mô đun phần mềm như sau.

### A. Cấu trúc phần cứng hệ thống

Cấu trúc phần cứng của hệ thống bao gồm hai mô đun chính: mô đun robot di động và mô đun điều khiển trung tâm. Các mô đun này được đặt cách xa nhau và giao tiếp với nhau thông qua mạng cục bộ không dây Wireless LAN (Hình 2).

Theo định nghĩa, *Robot di động* được đề cập đến như là một robot với cơ cấu bánh xe gắn dưới để có khả năng di chuyển, hoạt động tự quản trị, thường thực hiện chức năng dẫn đường hoặc thám hiểm [8]. Trong hệ thống an ninh tòa nhà, robot di động được sử dụng là robot Sputnik của hãng DrRobot [9] bao gồm các thành phần cơ bản sau: sáu cảm biến hồng ngoại gắn xung quanh robot; ba cảm biến siêu âm lần lượt gắn ở phía trước, bên phải và bên trái của robot; hai camera màu cho kích cỡ ảnh 353x288 với tốc độ lấy mẫu 15 khung hình/giây; hai cảm biến phát hiện sự chuyển động của người HMS (Human Motion Sensor) trong khoảng 150 cm; và bốn động cơ điều khiển điều khiển toàn bộ chuyển động của robot. Nguồn pin cho phép robot hoạt động trong khoảng 90 phút. Ngoài ra, robot có thể kết nối với bộ điều khiển trung tâm qua mạng cục bộ không dây Wireless LAN hoặc trực tiếp với mạng Internet qua mô đun không dây wi-fi 802.11. Hình 1 mô tả cấu trúc cơ khí và kết nối phần cứng của hệ robot tuần tra Sputnik.

Trong hệ thống robot di động tuần tra, chúng tôi sử dụng ba cảm biến siêu âm cho nhiệm vụ tránh vật và đi men theo tường (wall-following), một camera để truyền hình ảnh về trung tâm điều khiển và hai cảm biến HMS để phát hiện những chuyển động bất thường.

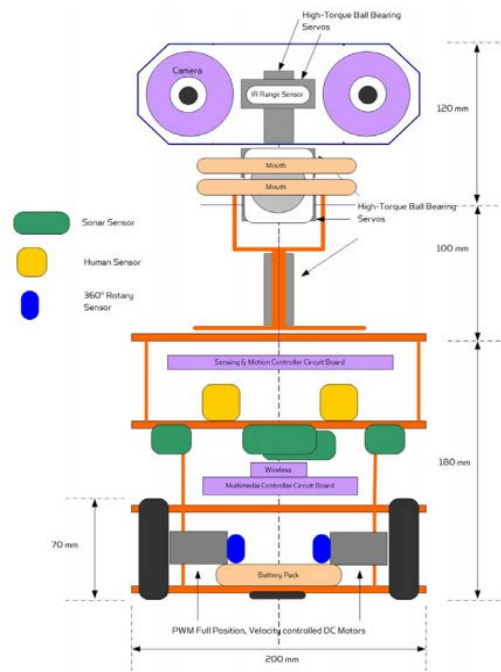

*Hình 1: Cấu trúc cơ khí của robot Sputnik*

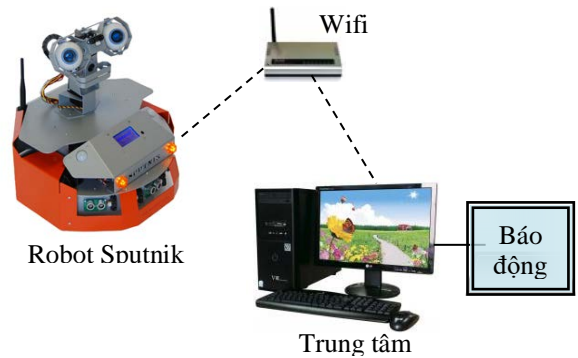

*Hình 2: Cấu trúc phần cứng hệ thống*

Mô đun điều khiển trung tâm bao gồm một máy chủ HP Intel Xeon 3.0Ghz chạy hệ điều hành Windows 7 64bits Professional. Mô đun này kết nối với robot thông qua mạng LAN không dây và kết nối với các phân hệ khác của tòa nhà thông qua hệ thống bus 485 [10]. Nhiệm vụ của mô đun là giám sát hoạt động của các phân hệ và đưa ra lệnh điều khiển ở mức ưu tiên cao trong những trường hợp khẩn cấp. Chi tiết về giao tiếp của mô đun trung tâm với các phân hệ đã được chúng tôi trình bày trong [1]. Trong bài báo này, mô đun trung tâm có thêm chức năng điều khiển robot và lưu trữ những hình ảnh do robot truyền về. Trường hợp có tín hiệu cảnh báo những xâm nhập bất hợp pháp, mô đun sẽ yêu cầu đóng tất cả các thiết bị vào ra tòa nhà đồng thời kích hoạt hệ thống báo động.

### B. Phần mềm điều khiển

Phầm mềm điều khiển được viết bằng ngôn ngữ Visual Basic và được cài đặt trên máy chủ HP của mô đun điều khiển trung tâm. Giao tiếp giữa chương trình phần mềm và robot được thực hiện qua các *giao diện lập trình ứng dụng API* (Application Programming Interface) với môi trường truyền dẫn là mạng LAN không dây sử dụng đa giao thức [11]. Như chỉ ra trong hình 3, phần mềm điều khiển bao gồm hai mô đun: mô đun xử lý trung tâm và mô đun logic mờ. Mô đun xử lý trung tâm thực hiện đồng thời các nhiệm vụ sau:

- Thu thập dữ liệu phản hồi của robot bao gồm: tốc độ và vị trí hiện tại, trạng thái nguồn pin Lithium, dữ liệu cảm biến hình ảnh, cảm biến siêu âm, cảm biến hồng ngoại và cảm biến phát hiện chuyển động HMS.
- Lưu trữ toàn bộ dữ liệu phản hồi và hiển thị tới giao diện người dùng GUI (Graphic User Interface).
- Thực hiện thuật toán wall-following xuất lệnh điều khiển chuyển động của robot .
- Gửi dữ liệu siêu âm tới khối logic mờ và nhận tín hiệu điều khiển phản hồi.
- Thực hiện lệnh báo động trong trường hợp nhận được yêu cầu từ robot.

Với các chức năng này, mô đun xử lý trung tâm đóng vai trò điều phối trung tâm và đảm bảo hoạt động nhịp nhàng của toàn hệ thống.

Mô đun logic mờ được phát triển dựa trên bộ công cụ MATLAB Fuzzy Logic Toolbox (MFLT) của Mathworks Inc [12]. Thuật toán logic mờ sau khi được cài đặt trên MFLT sẽ được đóng gói thành một đối tượng và nhúng trực tiếp vào trong mô đun điều khiển trung tâm viết bằng Visual Basic. Chi tiết việc phân tích và cài đặt thuật toán logic mờ được giới thiệu ở phần tiếp theo.

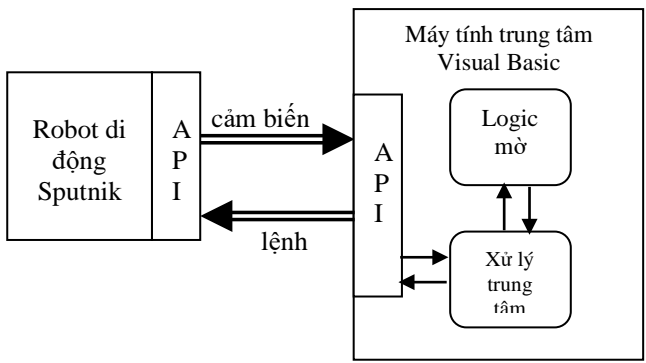

*Hình 3: Cấu trúc phần mềm hệ thống*

### III. THUẬT TOÁN ĐIỀU KHIỂN LOGIC MỜ

Áp dụng cho hệ thống robot di động tuần tra trong công trình này, lý thuyết logic mờ được ứng dụng để giải quyết vấn đề tránh vật cản. Khi đi tuần tra, robot xuất phát và đi bám tường dọc theo hành lang sử dụng thuật toán wall-following. Vì môi trường là không biết trước, nếu gặp vật cản, robot sẽ thực hiện việc tránh vật cục bộ và sau đó tiếp tục đi bám tường.

Thuật toán wall-following sử dụng cảm biến siêu âm phía bên trái của robot để điều khiển robot di chuyển dọc theo tường với một khoảng cách cố định là 30cm. Robot sẽ xuất phát theo hướng có cảm biến siêu âm bên trái gần với tường.

Khi đó, bức tường được coi như là đối tượng bám của robot và là vật cản của cảm biến. Cảm biến siêu âm sẽ đo khoảng cách từ robot tới bức tường, nếu giá trị này là 30cm thì điều khiển robot đi thẳng, nếu nhỏ hơn 30cm thì robot quay phải, lớn hơn 30cm thì quay trái. Do hàm điều khiển của robot Sputnik đối với trường hợp đi thẳng và quay góc là độc lập với nhau, nên để tránh trường hợp robot bị "kẹt" với một giá trị góc quay nào đó thì sau mỗi bước quay robot sẽ đi thẳng một đoạn 10cm rồi tiếp tục cập nhật giá trị cảm biến. Trong trường hợp để phát hiện sự chuyển động của người, hệ thống dựa vào hai cảm biến HMS.

Đối với yêu cầu tránh vật, hệ thống sẽ sử dụng hai cảm biến siêu âm ở giữa và bên phải cho thuật toán logic mờ. Bộ điều khiển mờ sử dụng các giá trị khoảng cách $U_2$ và $U_3$ của hai cảm biến siêu âm phía trước và bên phải là biến lối vào, biến lối ra là góc quay $\alpha$ mà robot cần thực hiện để tránh vật. Khi đó, robot sẽ tránh vật theo hướng phía trước và bên phải của robot. Theo cấu trúc cơ khí của robot, hai cảm biến siêu âm này sẽ tạo thành dải hoạt động trong phạm vi từ -20º đến +60º. Sơ đồ khối xác định các biến vào ra của bộ điều khiển mờ được thể hiện ở hình 4.

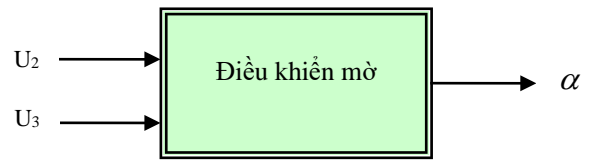

*Hình 4: Các biến vào ra*

Trong hệ thống của chúng tôi, việc cài đặt thuật toán logic mờ được thực hiện qua bốn bước bao gồm: xác định các biến ngôn ngữ vào ra, định nghĩa các hàm thuộc của các biến ngôn ngữ, thiết lập các luật mờ và cuối cùng là thực hiện việc giải mờ.

Các biến ngôn ngữ vào/ra được xác định như sau:
- $U_2, U_3$ : {Gần (G), Trung bình (TB), Xa (X)}
- $\alpha$ : {Âm ít (AI), Không (K), Dương ít (DI), Dương vừa (DV), Dương nhiều (DN)}

Giới hạn của cảm biến siêu âm từ 4-255 cm trong khi động tác tránh vật chỉ cần thực hiện tại những khoảng cách gần, vì thế hệ thống giới hạn phạm vi xử lý thông tin trong khoảng 4-100 cm và chia thành ba khoảng gần, trung bình, xa là đủ để robot có thể tránh vật an toàn. Góc quay robot thực hiện được trong phạm vi -20º đến +60º để phù hợp với phạm vi quét của cảm biến siêu âm.

Hàm thuộc của các biến ngôn ngữ được định nghĩa là hàm dạng hình thang và hình tam giác, những hàm cơ bản có khả năng tích hợp đơn giản (hình 5).

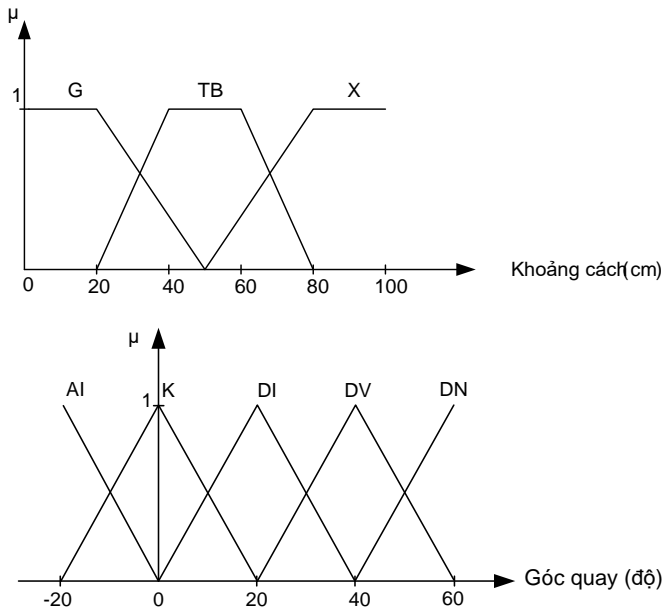

*Hình 5: Hàm thuộc của các biến vào/ra*

Luật mờ được xây dựng dựa trên mệnh đề suy diễn "Nếu….Thì…." và theo luật hợp thành Max-Min tạo thành 9 luật theo bảng 1 như sau:

|   |   | $U_3$ |   |   |
|---|---|---|---|---|
|   |   | **G** | **TB** | **X** |
| $U_2$ | **G** | DN | DI | DV |
|   | **TB** | AI | K | K |
|   | **X** | AI | K | K |

*Bảng 1: Luật mờ của các biến vào/ra*

Cuối cùng là bước giải mờ để xác định rõ giá trị góc quay lối ra. Phương pháp giải mờ được chọn là phương pháp điểm trọng tâm theo công thức:

$$\varphi = \frac{\sum x_i \mu(x_i)}{\sum \mu(x_i)} \quad (1)$$

trong đó, $x_i$ là giá trị trong miền xác định của tập mờ lối ra và $\mu(x_i)$ là giá trị của hàm thuộc của điểm $x_i$ trong miền đó. Đây là phương pháp được sử dụng nhiều nhất bởi khả năng cho phép xác định giá trị lối ra với sự tham gia của tất cả các tập mờ đầu ra của luật điều khiển một cách bình đẳng và chính xác.

Sau bước này, ta sẽ xác định được góc quay tránh vật của robot. Cũng giống trường hợp của thuật toán bám tường, vì hoạt động đi thẳng và quay góc là độc lập, nên để tránh bị "kẹt" thì robot sẽ đi thẳng một đoạn 10cm sau mỗi bước quay tránh vật. Lưu đồ thuật toán của toàn hệ thống được thể hiện ở hình 6.

## VI THỰC NGHIỆM

Một hệ thống robot tuần tra kiểm soát an ninh tòa nhà đã được chúng tôi xây dựng và cài đặt tại tầng 3 tòa nhà G2 thuộc Khoa Điện tử - Viễn thông, trường Đại học Công nghệ, Đại học Quốc gia Hà Nội. Nhiều thực nghiệm đã được tiến tại các thời điểm khác nhau, trong những điều kiện khác nhau và với nhiều tình huống giả định có thể xảy ra. Mục đích của các thực nghiệm là kiểm tra tính ổn định và hiệu quả của hệ thống để từ đó có đánh giá về tính khả thi của giải pháp đề ra đối với yêu cầu an ninh tòa nhà.

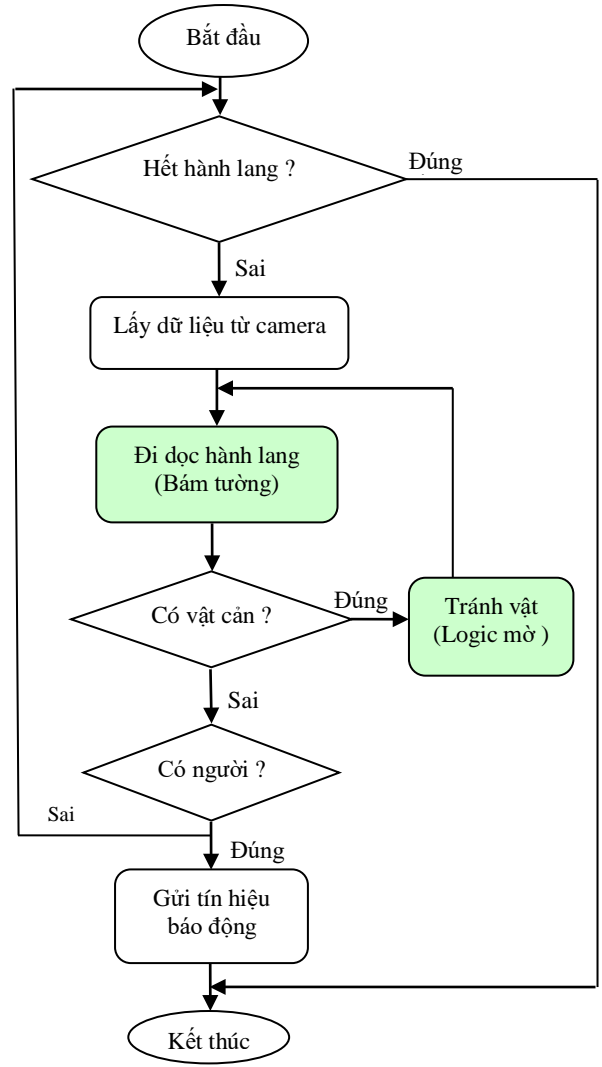

*Hình 6: Lưu đồ thuật toán điều khiển*

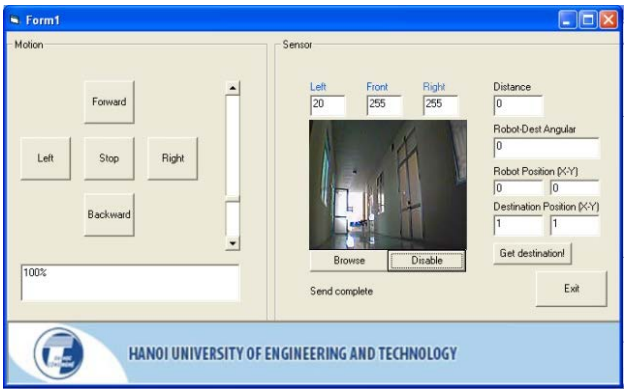
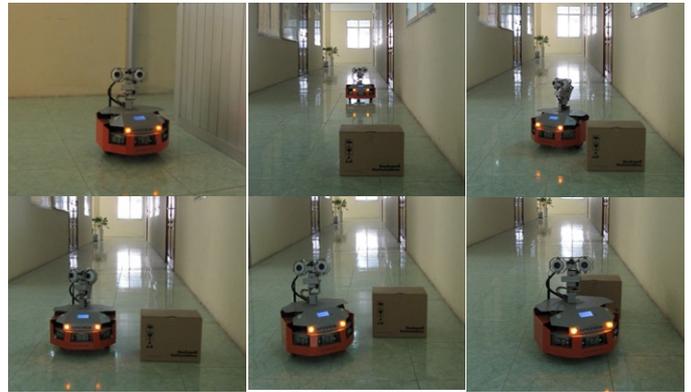

*Hình 7: Giao diện người sử dụng*  *Hình 8: Hình ảnh robot trong quá trình tuần tra*

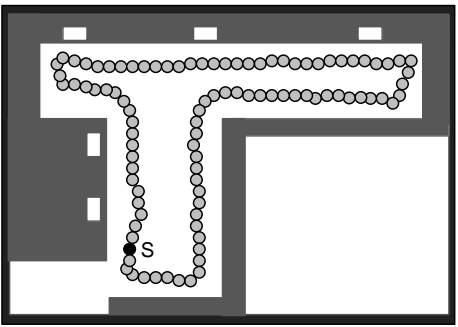
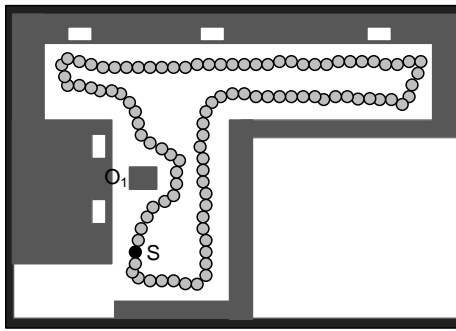
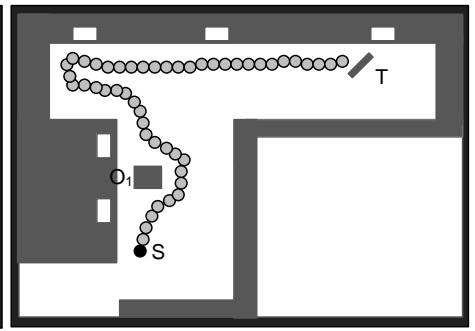

*Hình 9: Robot di chuyển dọc hành lang*  *Hình 10: Robot di chuyển tránh vật cản*  *Hình 11: Robot phát hiện chuyển động của người*

Hoạt động tuần tra của robot được thực hiện tại hành lang nhà G2 với cấu trúc được trình bày như trong hình 9. Tổng chiều dài một vòng quanh hành lang là 5000cm, chiều rộng 220cm. Trong các thực nghiệm của chúng tôi, robot di chuyển với vận tốc trung bình 10cm/s. Như vậy, với thời gian nguồn pin cho phép hoạt động 90 phút, robot có thể thực hiện được 10 lần tuần tra trong mỗi phiên làm việc. Nếu đặt thời gian tuần tra là từ 8h tối tới 6h sáng, robot sẽ thực hiện được trung bình 1 lần tuần tra/giờ. Giá trị này là phù hợp cho hoạt động đảm bảo an ninh tòa nhà.

Ở thực nghiệm thứ nhất, chúng tôi kiểm tra hoạt động bám tường của robot sử dụng thuật toán wall-following. Robot xuất phát tại điểm S, đi dọc theo hành lang không gặp vật cản nào và trở về điểm S ban đầu. Để nhận ra điểm S, robot sử dụng một khoảng trống trước điểm này. Khi giá trị dữ liệu từ cảm biến nhận được là 255 (giá trị không có vật cản) trong 5 lần đọc liên tiếp, robot sẽ quay một góc $90^o$ và đi thẳng một đoạn 10cm để kết thúc tại điểm S. Hình 9 trình bày đường đi của robot được ghi lại. Trong thực nghiệm này, robot đã hoàn thành một vòng tuần tra với thời gian hoạt động là 9 phút.

Hình 10 biểu diễn đường đi của robot trong trường hợp có một vật cản được đặt tại điểm $O_1$ trên hành lang. Khi đó, dựa vào thuật toán logic mờ, robot di chuyển tránh vật và tiếp tục hoạt động tuần tra dọc theo hành lang. Hình 12 biểu diễn quan hệ giữa góc quay lối ra tránh vật theo giá trị của hai cảm biến siêu âm lối vào $U_2$, $U_3$. Từ hình vẽ, chúng ta thấy với các giá trị của hai cảm biến lớn, tương ứng với trường hợp không phát hiện vật cản, thì giá trị của góc quay là $0^0$, nghĩa là robot sẽ đi thẳng bám tường. Trong trường hợp có vật cản, giá của các cảm biến nhỏ, giá trị góc quay sẽ thay đổi khác $0^0$ dựa trên thuật toán logic mờ để robot thực hiện việc tránh vật.

Trong thực nghiệm tiếp theo, robot xuất phát từ điểm S đến điểm T thì cảm biến HMS phát hiện ra chuyển động của người, khi đó robot đã gửi một tín hiệu về trung tâm để trung tâm kích hoạt hệ thống báo động và dừng hoạt động tuần tra (hình 11). Trong tất cả các trường hợp thử nghiệm trên, khi di chuyển, robot liên tục gửi dữ liệu từ các cảm biến và camera với tốc độ 15 khung hình/giây về trung tâm điều khiển. Nhân viên an ninh có thể ngồi tại máy tính trung tâm để giám sát tình trạng an ninh tòa nhà thông qua dữ liệu gửi về từ robot và có thể điều chỉnh vị trí của camera để có góc nhìn thích hợp. Ngoài ra, robot cũng được trang bị cơ chế điều khiển bằng tay. Do đó, nhân viên an ninh có thể trực tiếp điều khiển robot khi cần thông qua giao diện chương trình điều khiển GUI (hình 7). Trong thực nghiệm của chúng tôi, máy chủ điều khiển trung tâm HP được đặt tại phòng thí nghiệm Điều khiển tự động và Robotics, phòng 315 tòa nhà G2. Hình 8 trình bày một số hình ảnh của robot trong quá trình tuần tra và tránh vật.

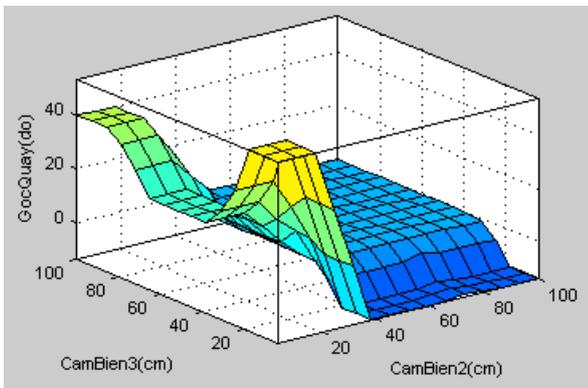

*Hình 12: Giá trị góc quay robot theo giá trị vào của hai cảm biến siêu âm thu được từ thuật toán logic mờ*

Trong quá trình tiến hành thực nghiệm, cũng có một số trường hợp robot đã không thành công trong hoạt động tuần tra. Đó là khi xuất hiện vật cản có góc cạnh nhỏ, vật cản đặt bên phải robot quá gần tạo thành hành lang hẹp, hay vật cản xuất hiện quá gần cả phía trước và phía bên phải robot. Nguyên nhân là do hạn chế về mặt cấu tạo của robot và đặc điểm của cảm biến siêu âm. Phạm vi quét của cảm biến siêu âm theo dạng hình quạt, diện tích bề mặt quét càng rộng thì càng dễ phát hiện, trong những trường hợp góc cạnh của vật cản quá nhỏ thì không phát hiện được. Ngoài ra, do robot Sputnik chỉ có ba cảm biến siêu âm gắn phía trước, bên trái và bên phải tạo thành góc quét hẹp trong phạm vi từ $-60^0$ đến $+60^0$ nên khả năng tránh vật bị hạn chế. Những hạn chế trên có thể được khắc phục khi lắp thêm các cảm biến siêu âm quanh thân robot hoặc kết hợp với các cảm biến khác như hồng ngoại để phân vùng tránh vật từ xa.

## V KẾT LUẬN

Trong bài báo này, chúng tôi đã đề xuất một giải pháp mới trong việc kiểm soát an ninh của hệ thống tòa nhà thông minh: sử dụng robot tuần tra có trang bị nhiều loại cảm biến để thay thế cho hệ thống nhiều camera truyền thống. Một hệ thống robot di động tuần tra đã được xây dựng thành công với các chức năng: tuần tra một hành lang của tòa nhà, tránh vật trong quá trình tuần tra, gửi dữ liệu cảm biến về trung tâm và kích hoạt hệ thống báo động khi phát hiện những chuyển động bất thường. Các kết quả tính toán và thực nghiệm đã chứng minh tính ổn định, hiệu quả và khả thi của hệ thống.

Trong thời gian tiếp theo, nhiều cảm biến siêu âm sẽ được trang bị thêm cho robot và thuật toán logic mờ sẽ được cải tiến để khắc phục những hạn chế còn tồn tại của hệ thống. Chúng tôi cũng sẽ tiến hành nghiên cứu và thực nghiệm với hệ thống gồm nhiều robot hoạt động đồng thời (multi-agent system) để đánh giá sự tương tác giữa các robot trong nhiệm vụ nâng cao hiệu quả kiểm soát an ninh.